\documentclass[10pt,twocolumn,letterpaper]{article}

\usepackage{iccv}
\usepackage{times}
\usepackage{epsfig}
\usepackage{graphicx}
\usepackage{amsmath}
\usepackage{amssymb}
\usepackage{multirow}
\usepackage{graphicx}
\usepackage{booktabs}
\usepackage{adjustbox}
\usepackage{mathtools}
\usepackage{authblk}
\usepackage{url}

\DeclarePairedDelimiter{\norm}{\lVert}{\rVert}
\usepackage[pagebackref=true,breaklinks=true,letterpaper=true,colorlinks,bookmarks=false]{hyperref}

\iccvfinalcopy % *** Uncomment this line for the final submission

\def\iccvPaperID{****} % *** Enter the ICCV Paper ID here

\begin{document}

%%%%%%%%% TITLE
\title{Adaptive Context Network for Scene Parsing}
\author[${1,4}$ ]{
Jun Fu
}
\author[${1}$]{
\; Jing Liu\thanks{Corresponding Author}
}
\author[${1}$]{
\; Yuhang Wang  
}

\author[${2}$]{
\\ Yong Li 
}

\author[${2}$]{
\ Yongjun Bao
}
\author[${3}$]{
\;Jinhui Tang
}
\author[${1}$]{
\;Hanqing Lu
}
\affil[ ]{$^1$National Laboratory of Pattern Recognition, Institute of Automation, Chinese Academy of Sciences\; $^2 $Business Growth BU, JD.com\; $^3$ Nanjing University of Science and Technology\;}
\affil[ ]{$^4$ University of Chinese Academy of Sciences\; }
\affil[ ]{ 
\tt\small \{jun.fu,jliu,luhq\}@nlpr.ia.ac.cn,wangyuhang.casia@gmail.com\\ \{liyong5,baoyongjun\}@jd.com,jinhuitang@njust.edu.cn }

\renewcommand\Authsep{  } 
\renewcommand\Authands{  }

\maketitle
\thispagestyle{empty}
%%%%%%%%% ABSTRACT
\begin{abstract}
   Recent works attempt to improve scene parsing performance by exploring different levels of contexts, and typically train a well-designed convolutional network to exploit useful contexts across all pixels equally. However, in this paper, we find that the context demands are varying from different pixels or regions in each image. Based on this observation, we propose an Adaptive Context Network (ACNet) to capture the pixel-aware contexts by a competitive fusion of global context and local context according to different per-pixel demands. Specifically, when given a pixel, the global context demand is measured by the similarity between the global feature and its local feature, whose reverse value can be used to measure the local context demand. We model  the two demand measurements by the proposed global context module and local context module, respectively, to generate adaptive contextual features. Furthermore, we  import multiple such modules to build several adaptive context blocks in  different levels of network to obtain a coarse-to-fine result.  Finally, comprehensive experimental evaluations demonstrate the effectiveness of the proposed ACNet, and new state-of-the-arts performances are achieved on all four public datasets, i.e. Cityscapes, ADE20K, PASCAL Context, and COCO Stuff.
\end{abstract}

%%%%%%%%% BODY TEXT
\section{Introduction}

Scene parsing is a fundamental image understanding task which aims to perform per-pixel categorizations for a given scene image. Most recent approaches for scene parsing are based on Fully Convolutional Networks (FCNs) \cite{FCN}. However, there are two limitations in FCN framworks. First, the consecutive subsampling operations like pooling and convolution striding lead to a significant decrease of the initial image resolution and make the loss of spatial details for scene parsing. Second, due to the limited receptive field \cite{parsenet,luo2016understanding} or local context features, the per-pixel dense classification is often ambiguous. In the end, FCNs result in the problems of rough object boundaries, ignorance of small objects, and misclassification of big objects and stuff.

\begin{figure}[!t]
        \centering
        \includegraphics[width=1\linewidth]{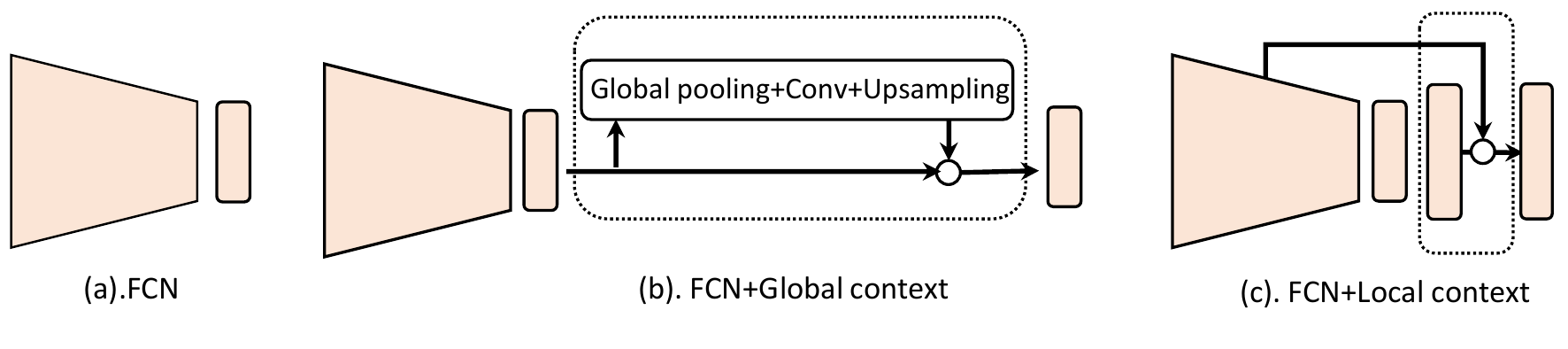}
       \includegraphics[width=1\linewidth]{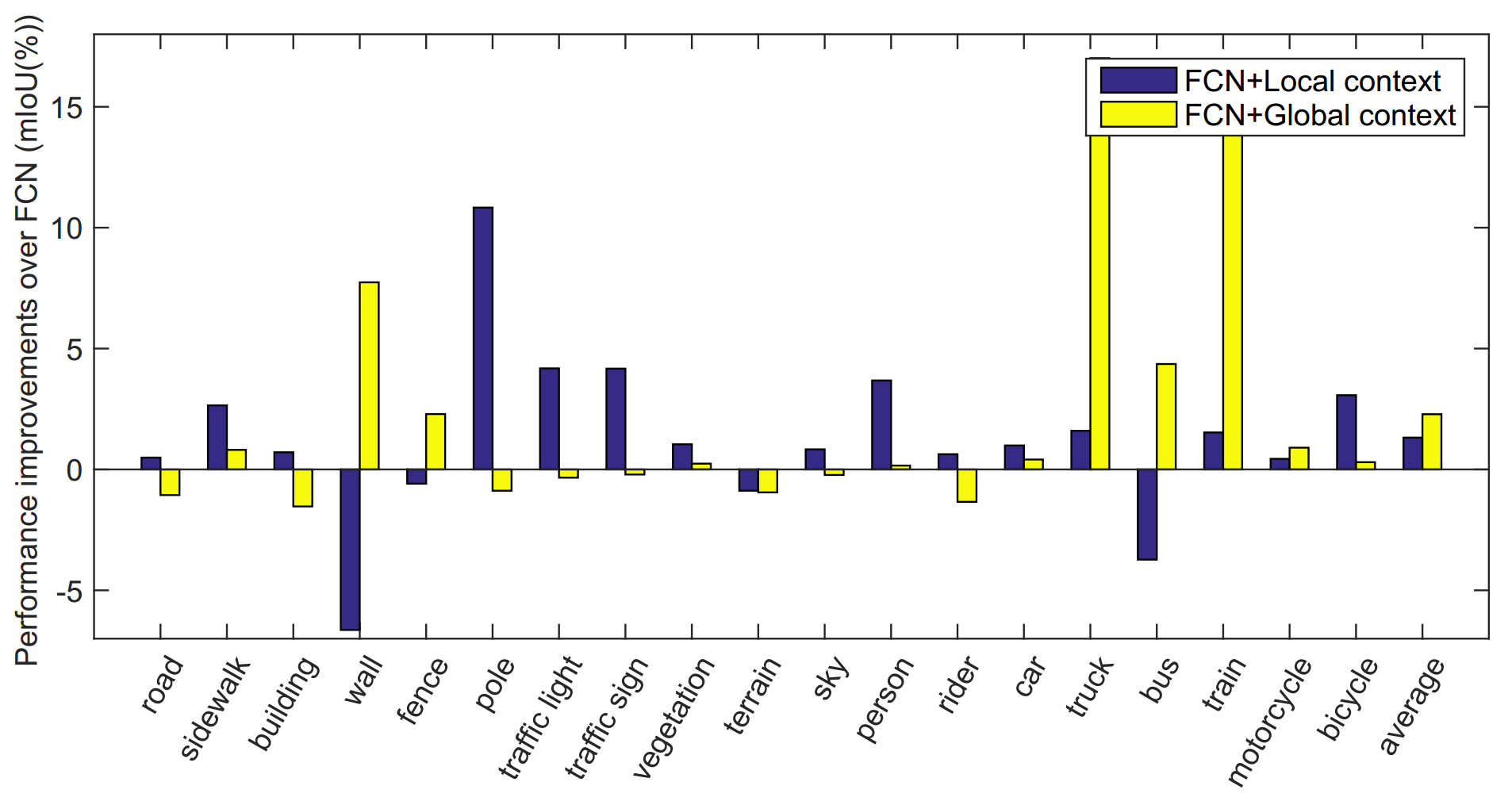}

        \caption{The performance improvements over the basic FCN (a. Dilated FCN) on Cityscapes val set with the help of global context (b. Dilated FCN+Global context) and local context (c. Dilated FCN+Local context). Specially, pixel-wise enhanced representation by the global average pooling feature are employed as the global context, and a concatenated representation with low-level features as the local context.  }
        \label{example}%
        \vspace{-1em}
\end{figure}

Throughout various FCN-based improvements to overcome the above limitations, effective strategies to utilize different levels of contexts (i.e., local context and global context) are the main directions. Specifically, some methods \cite{refinenet,exfuse, yu2018learning,fu2017stacked} adopt ``U-net'' architectures, which exploit multi-scale local contexts from middle-layers, to complement more visual details. Some methods \cite{deeplabv2,yu2015multi} employ dilated convolution layers to capture a wider context with a larger receptive field while maintaining the resolution. Besides, the image-level features obtained by global average pooling \cite{parsenet,deeplabv3,pspnet} are proposed  as a global context to clarify local confusion. However, these FCN-based variants adopt per-pixel unified processing and overlook different per-pixel demands on different levels of contexts.  
That is, the local context from middle layers is essential for the class prediction of those pixels on edges or small objects, while the global context exploring the image-level representation is benefit to categorize large objects or stuff regions, especially for the case when the target region exceeds the receptive field of the network.
We can also observe the necessity of pixel-sensitive context modeling from the results comparison shown in Figure \ref{example}, in which the local and global contexts achieve different improvements on different objects or stuff. 
Therefore, how to effectively capture such pixel- or region-aware contexts in an end-to-end training framework is an open but valuable research topic for comprehensively accurate scene parsing. 

In this paper, we propose an Adaptive Context Network (ACNet) to capture the pixel-aware contexts for image scene parsing. Different from previous methods which fuse different-level contexts for each pixel equally,
ACNet generates different per-pixel contexts, i.e., the context-based features are functions of the input data and also vary from different pixels. Such an adaptive context generation is achieved by a competitive fusion mechanism of the global context from image-level feature and local context from the middle-layer feature according to per-pixel different demands. In other words, with the more attention paid on global context for a certain pixel, the less attention is paid on local context, and vice versa. 

Usually, the global average pooled feature has a semantic guidance for large objects and stuff, but it lacks spatial information which makes it different from the features of details. Hence, we can match the global pooled feature with the feature of each pixel, obtaining the possibility of the pixel to be an element of large objects or spatial details. It can be further used as a pixel-aware context guidance to adaptively fuse the global features (global context) and low-level features (local context). 
Motivated by this intuition, we propose a \emph{global context module} to adaptively capture global context.  By measuring the similarity between the global feature and per-pixel feature, we can obtain the pixel-aware demanding extent, called as global gated coefficient. The larger gated coefficient indicates that more global context and the less local context could be fused to the pixel. Then we multiply the global feature with the pixel-aware global gated coefficient before adding it to the pixel feature, with which some mislabeling and inconsistent results can be further corrected.

We also propose a \emph{local context module} to compensate spatial details according to the local context demands.
Specifically, we find that the pixels with features dissimilar to global feature, trend to be detail parts of images and need more local context to obtain precise results. Hence, we regard the reverse value of the  global gated coefficient as local gated coefficient and  multiply it with the low-level feature to generate a  local gated feature. It emphasizes pixel-aware local context to spatial details and avoids some noises to the pixels belonging to big objects. Furthermore, we reuse multiple  local gated features, which is similar to a recurrent learning process and complements more detail information.

We jointly employ a global context module and a local context module as an adaptive context block, and import such blocks into different levels of network.  The architecture of our proposed ACNet is shown in Figure \ref{CAN}.  Finally, comprehensive experimental analyses on Cityscapes dataset \cite{cityscapes}, ADE20k\cite{zhou2017scene}, PASCAL Context \cite{pcontext}, and COCO stuff\cite{caesar2018coco} dataset demonstrate the effectiveness of ACNet.

The main contributions of this paper are as follows:

\begin{itemize}
   \item We propose an Adaptive Context Network (ACNet) to improve contextual information fusion according to the context demands of different pixels.

   \item A novel mechanism  is proposed to measure  global context demand. Global pooled feature can be  adaptively fused to the pixels which need large context, thus reducing misclassification for large objects or stuff. 

   \item We improve  local context fusion according to the local context demand and reuse local feature progressively, thus improving segmentation results on small objects and edges. 

   \item ACNet achieves new state-of-the-art performance on various scene parsing datasets. In particular, our ACNet achieves a Mean IoU score of 82.3\% on Cityscapes testing set without using coarse data, and 45.90\% on ADE20K validation set, respectively.

\end{itemize}

\section{Related Work}
\noindent{\bfseries Global context embedding.} Global context embedding have been proven its effectiveness to improve the categorization of some large semantic regions. ParseNet \cite{parsenet} employs  the global average pooled feature to augment the features at each location.  
PSPNet \cite{pspnet} applies the global average pooling in their Spatial Pyramid Pooling module to collect global context.  The work \cite{hung2017scene} captures  global contexts by a gloabl context network based on scene similarities.
 BiSeNet  \cite{yu2018bisenet} adds the global pooling on the top of the encoder structure to capture the global context. EncNet \cite{encnet} employs an encoding layer to capture global context and selectively highlight the class-dependent featuremaps.

\noindent{\bfseries Local context embedding.}  U-net based methods often adopt local context from  low- and middle-level visual features to generate sharp boundaries or small details for high-resolution prediction.
RefineNet \cite{refinenet}  utilizes an encoder-decoder framework and refines low-resolution segmentation with fine-gained low-level feature. ExFuse \cite{exfuse}  assigns auxiliary supervisions directly to the early stages of the encoder network for improving low-level context. Deeplabv3+ \cite{chen2018encoder} 
adds a simple  decoder module to capture local context, refining the segmentation results.

\noindent{\bfseries Attention and gating mechanisms.} Attention mechanisms have been widely used to  improve the performance of segmentation task. PAN \cite{li2018pyramid} uses a global pooling to generate global attention, which can select the channel maps effectively. LRR \cite{ghiasi2016laplacian}  generates a multiplicative gating to refine segment boundaries reconstructed from lower-resolution score maps.  Ding et al.\cite{ding2018context}  proposes a scheme of RNN-based gating mechanism  to selectively aggregate multi-scale score maps, which can achieve an optimal multi-scale aggregation. The works \cite{fu2018dual,yuan2018ocnet,huang2018ccnet} adopt self-attention mechanism to model the relationship of features.

Different from these works,we introduce a data-driven gating mechanism to capture global context and local context according to pixel-aware context demand.

\section{Adaptive Context Network}

\subsection{Overview}
Contextual information is effective for scene parsing task, most of current methods  fuse the different context  to each pixel equally, ignoring the different demands of  pixel-aware contexts.
In this work, we propose a novel Adaptive Context Network (ACNet) to weigh the global and local context complemented to each pixel by a competitive fusion mechanism.

\begin{figure}[!t]
        \centering
        \includegraphics[width=1\linewidth]{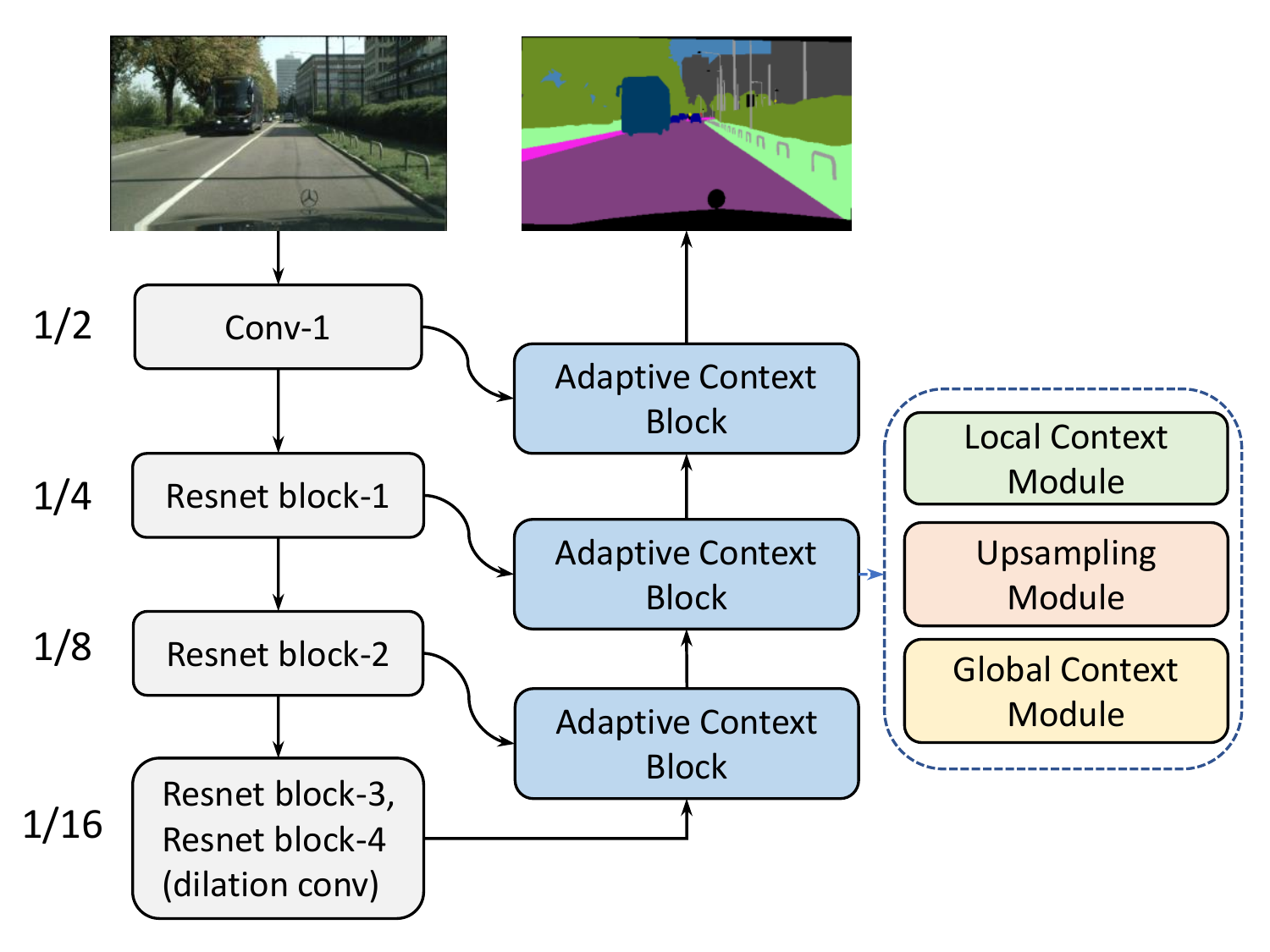}
        \vspace{-1em}
        \caption{Overview of Adaptive Context Network. (Best viewed in color)}
         \label{CAN}%
        \vspace{-0.5em}
\end{figure}

The overall architecture is shown in Figure \ref{CAN}, which adopts pretrained dilated ResNet \cite{he2016deep}, as the backbone network, and  multiple adaptive context blocks to progressively generate high-resolution segmentation map.
In the backbone network, we  remove the  downsampling operations and employ dilated convolutions in the last ResNet blocks, thus obtaining  dense feature  with output size 1/16 of the input image. It could achieve the balance between retaining spatial details and  computation cost \cite{chen2018encoder}. 
In upsampling process, three adaptive context blocks are employed with three different resolutions.
Each adaptive context block consists of a global context module, an upsampling module and a  local context module, where the global context module selectively  captures global context from the high-level features and the local context module selectively  captures local context from the  low-level features.

In the following subsections, we will elaborate the designing details of the global context module, the local context module and their aggregation within an adaptive context block.

\begin{figure*}[!t]
        \centering
        \includegraphics[width=1\linewidth]{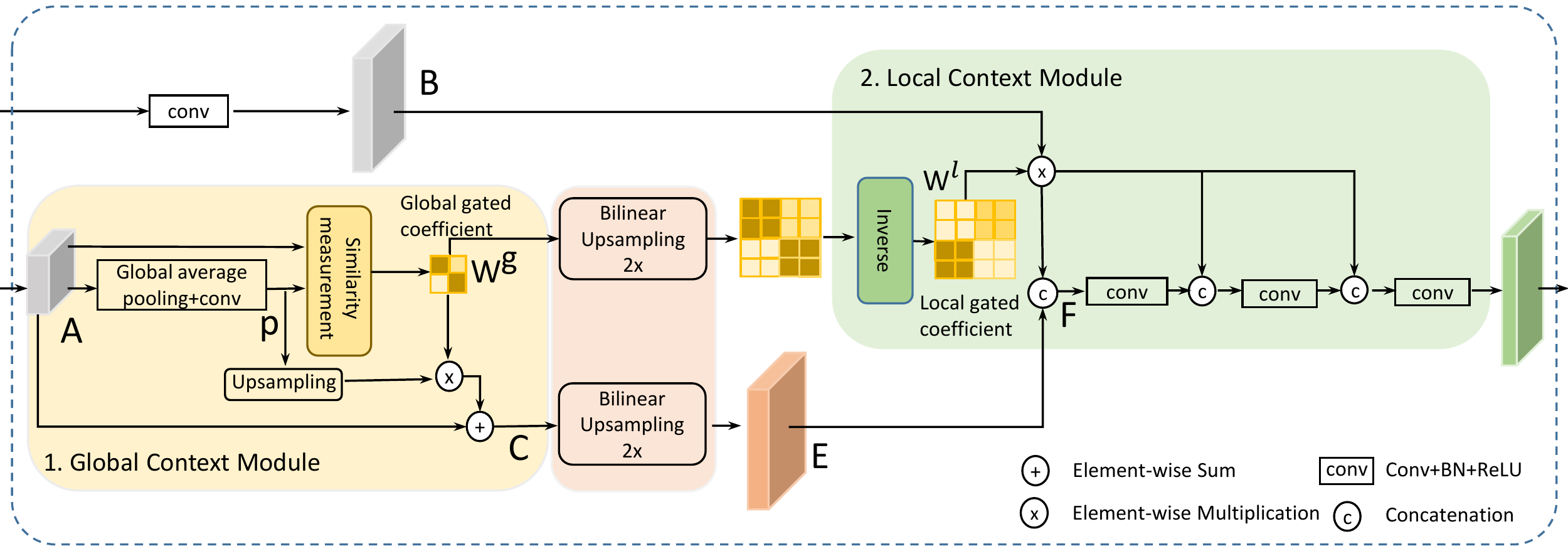}
        \caption{The details of Adaptive Context Block including (1) Global Context Module and (2)  Local Context Module. (Best viewed in color)}
         \label{ACB}%
        \vspace{-1em}
\end{figure*}

\subsection{Global Context Module}
Global context can provide global semantic guidance for overall scene images, thus rectifying misclassification and inconsistent parsing results. However, the benefit from global context is different for large objects and spatial details. 
it is necessary to treat each pixel differently when exploring the global context, that is to say, some pixels need more global context for  categorization, while others may do not.
Based on the intuition that the global pooled feature prefers to the large objects and stuff and lacks spatial information, we can match the global feature with the feature in each pixel,  obtaining its possibility to be as an element of large objects or spatial details.   
Then we can exploit it to adaptively fuse the global context.
To this end, we propose a Global Context Module (GCM) as follows.

Given an input feature map $\mathbf{A} \in \mathbb{R}^{C \times H \times W}$, we use a global average pooling following a convolution layer to  generate a global feature $p \in \mathbb{R}^{C \times 1 \times 1}$. 
In order to obtain the pixel-aware demand for global context (global gated coefficient), we first measure the feature similarity  by calculating the  euclidean distance  $\mathbf{D} \in \mathbb{R}^{H \times W}$ between global feature $p$ and  the features ${a_{i} \in \mathbf{A}}$ for each pixel $i$, ${d_{i} \in \mathbf{D}}$ is denoted as:
\begin{equation}
d_{i}=\norm{a_{i}-p}_{2}
\end{equation}
where ${a_{i} \in \mathbf{A}}$ , $i \in [1,2,...,H\times W]$ is  ${i^{th}}$ location in $ \mathbf{A}$.
Noted that  the smaller $d_{i}$ indicates that the feature at ${i^{th}}$ locations is closer to the global feature.  
Then we generate a global gated coefficient $\mathbf{W^g} \in \mathbb{R}^{H \times W}$  which is smoothed by an exponential function, ${w^g_{i} \in \mathbf{W^g}}$ is denoted as:
\begin{equation}
{w^g_{i}}=exp(-\frac{d_{i}-k }{ \delta} )
\end{equation}
where  $k$  is set to $\min_{i=1}^ {H\times W} (d_{i})$  for limiting the range of $w^g_{i} \in (0,1]$. And $\delta$ is a hyperparameter, which controls the amplitude of the difference between high response and low response.

Finally, we  multiply  the global feature $p$ by $w^g_{i}$ and a scale parameter $ \alpha $, and  then perform an element-wise sum operation with the features $\mathbf{A}$ to obtain the final output $\mathbf{C}\in \mathbb{R}^{C \times H \times W}$, ${c_{i} \in \mathbf{C}}$ is denoted as:

\begin{equation}
c_{i} =  \alpha w^g_{i}p + a_{i}
\label{equ1}%
\end{equation}
where $\alpha$ is a learned factor  and  initialized as 1. Here, we adopt sum operation instead of concatenation for saving memory. The details of global context module is shown in Figure.\ref{ACB} (1).

 It can be inferred from the above formulation that the feature $\mathbf{C}$  at different position obtain different global context according to global gated  coefficient $\mathbf{W^g}$. 

With this design, GCM could  selectively enhance semantic consistency and  reduce the  misclassification and inconsistent predictions for large objects or stuff.

\subsection{Local Context Module}
Local context contributes to refine object boundaries and details. However, many methods fuse local context  to  all pixels without considering the different demand for local context.
To solve this problem, we propose
a Local Context Module (LCM) to selectively fuse local context for better refined segmentation.

As mentioned  in Section 3.2,  the  global gated coefficient with high response indicates the pixels belong to large objects and stuff while low response indicates the pixels belong to spatial details. Based on this observation, we could obtain the local gated coeffcient by reversing value of global gated coefficient, where the global gated coefficient have been upsampled, formulated as:
\begin{equation}
\mathbf{W^l} =\mathbf{1}-up(\mathbf{W^g})
\end{equation}
where $up$($\cdot$) denotes a bilinear interpolation operation.
In this way, the local gated coefficient indicates the more possibility the pixels belong to spatial details, the more local contexts are required, and vice versa. Then we obtain pixel-aware local context (gated local features) by multiply the local feature $\mathbf{B}\in \mathbb{R}^{C \times H \times W}$ from middle-layer features with  the local gated coefficient and a scale parameter $\beta $. Finally, we concat the feature with the upsampled feature $\mathbf{E}\in \mathbb{R}^{C \times H \times W}$ to generate a refined feature $\mathbf{F}\in \mathbb{R}^{C \times H \times W}$, ${f_{i} \in \mathbf{F}}$ is denoted as:

\begin{equation}
f_{i} =  cat( \beta w^l_{i}b_{i},e_{i})
\label{equ1}%
\end{equation}
where $cat$($\cdot$) denotes a concatenation operation, and $\beta$ is a learned factor  and  initialized as 1. We adopt concatenation operation to combine the gated local feature and high-level feature, and  a convolution layer is employed to fuse them. The details of local context module is shown in Figure.\ref{ACB}(2).
% 解说公式
With this design, we can selectively aggregate the local context according to the  context demand of each pixel.

In addition, we find that it is useful to introduce gated local context directly multiple times. Specifically, we reuse gated local features by  
a concatenation operation followed by  a convolution layer for three times.
Such a recurrent learning process complements more spatial details for each position, and achieve a coarse-to-fine performance improvement.  Noted that, it haven't been discussed in previous works \cite{refinenet, exfuse,yu2015multi,chen2018encoder}. And we also  verify the effectiveness of this process in experiments.

\subsection{Adaptive Context Block}
Based on GCM and LCM, we further design an Adaptive Context Block to  selectively  capture global and local contextual information simultaneously.

Adaptive context block is built upon a cascaded architecture, the  high-layer features are first fed into  a global context module to  selectively fuse global context to each pixel. 
Then  passed sequentially through a bilinear upsampling layer and   a local context module for learning a restoration of refined features. In order to obtain resolution corresponding to low-level feature, we also enlarge spatial resolution of the global gated coefficient by a bilinear upsampling operation before feeding into the local context module. Following \cite{chen2018encoder}, we apply a convolution layer on the low-level features to reduce the number of channels, thus refining the  low-level features. 

In the adaptive context block, we introduce a competitive fusion mechanism to capture global and local  context  according to their correlation of gated coefficient, thus suitable context can be adaptively fused to each pixels for better feature representation.

\section{Experiments}
The proposed method are evaluated on Cityscapes \cite{cityscapes}, ADE20K \cite{zhou2017scene}, PASCAL Context \cite{pcontext}, COCO Stuff \cite{caesar2018coco}.
Experimental results demonstrate that ACNet achieves new state-of-the-art performance on these  datasets. 
 In the next subsections,  we first introduce the datasets and  implementation details, then we make detail comparisons to evaluate our approaches on Cityscapes dataset. Finally, we present our  results  compared with state-of-the-art methods on ADE20K, PASCAL Context, COCO Stuff dataset.

\subsection{Datasets}
\noindent{\bfseries Cityscapes} The dataset is a well-known road scene dateset collected for scene parsing, which
 has  2,979 images for training, 500 images  for validation and 1,525 images for testing. 
Each image has a high resolution of ${2048\times1024}$ pixels with 19 semantic classes. Noted that no coarse data is employed  in our experiments.
 
\noindent{\bfseries ADE20K} The ADE20K dataset is a vary challenge scene understanding dataset, which  contains 150 classes (35 stuff classes and 115 discrete object classes). The dataset is divided into 20, 210/2, 000/3, 352 images for training, validation and testing. 

\noindent{\bfseries PASCAL Context} The dataset is widely used for scene parsing, which contains 4,998 images for training and 5,105 images for testing. Following previous works \cite{refinenet,encnet}, we evaluate the method on  60 categories ( 59
classes  and one background category ).

\noindent{\bfseries COCO Stuff}  The dataset has 171 categories including 80 objects and 91 stuff
annotated to each pixel.  Following previous works \cite{ding2018context,DAGRNN,refinenet}, we adopt 9,000 images for training and 1,000 images for testing.

\subsection{Implementation Details}

 We employ a dilated pretrained ResNet architecture as our backbone network, where the dilated rates in the last ResNet block is set to (2,2,2). 
Following \cite{encnet,pspnet}, we  apply a $3 \times3$ convolution layer with BN, ReLU on the outputs of the last ResNet block to reduce the number of channels to 512 before feeding into the first adaptive context block.
In addition, we adopt the outputs of ResNet block-1 and ResNet block-2 as the low-level features, which provide local context for  the first two adaptive context blocks. And we only adopt a global context module in last adaptive context block. In the first two adaptive context block,we employ a $3 \times3$ convolution layer on  the low-level featues before feeding it into local context module. The other convolution layers in the first two adaptive context block are composed of a $3 \times3$ convolution operation with 448 and 256 kernels respectively followed by BN and ReLU. Pytorch is used to implement our method.

During training phase, we employ a ‘poly’ learning rate policy where
the initial learning rate is multiplied by $(1-\frac{iter}{total\_iter})^{0.9}$ after each iteration, and enable synchronized batch normalization \cite{encnet}. The base learning rate is set to 0.005 for Cityscapes and  ADE20K, 0.001  for PASCAL Context and COCO stuff. Momentum and weight decay  coefficients are set to 0.9 and 0.0001 respectively. Following \cite{pspnet}, auxiliary loss is adopted when we adopt the bockbone ResNet101.
In addition, we apply  random cropping and random left-right flipping during training phase, and the randomly scaling for data  augmentation is not employed if not mentioned on Cityscapes dataset.

\subsection{Results on Cityscape dataset}

\begin{figure*}[!t]
        \centering
        \includegraphics[width=1\linewidth]{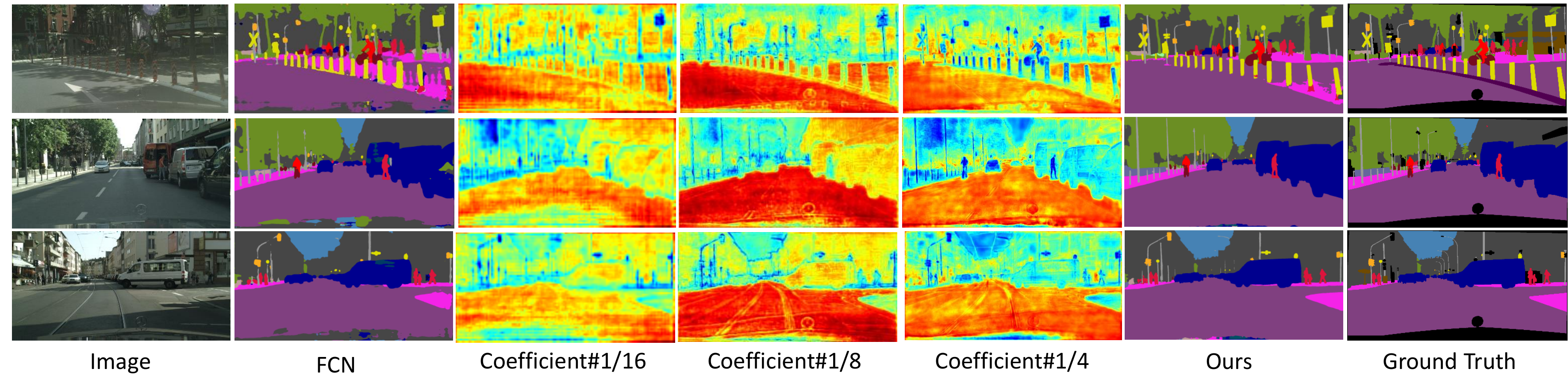}
        \caption{Visualization results of global gated coefficient from
global context module with 1/16, 1/8 and 1/4 resolution respectively, we can find that the  pixels with large global gated coefficient perfer to dominant stuff and large objects. Compared with FCN, our method enhance semantic guidance with global context in the regions with large coefficient and provide more local context in other regions, thus obtaining accurate segmentation results. (Best viewed in color) }
         \label{attencity}%
      %  \vspace{-1em}
\end{figure*}

\noindent{\bfseries Global Context Module:} Firts of all, we design a global context module to adaptively aggregate global context according to pixel-aware demands. 
Specifically, we follow \cite{deeplabv2} and build two dilated networks (ResNet-50) which yield the final feature maps with the 1/8 and 1/16 size of the original image. Next, the global context are added on the top of the networks with two different settings, which are  GC and GCM respectively (GC denotes that we directly sum the global features to each pixel equally, GCM represents  the global context module).

Experimental results are shown in Table. \ref{globalT}, we can see that GCM (global context module) achieves better performance than GC in both two settings,  especially for the output 1/8 size of the original image. It shows the effectiveness of GCM and also indicates that the improvement will be more obvious if the higher-resolution  global gated coefficient is produced in GCM based on the dilated FCN.
In addition, we also provide a discussion about $\delta$, which controls the amplitude of the difference between high response and low response of the global gated coefficient (mentioned in Sec.3.1). When we set $\delta$ to 5, the gloabl context module yields the best performance. We fix this value and employ  the lowest resolution output 1/16 size of the original image in following experiments.

\begin{table}[t]
\begin{center}
\begin{tabular}{l|c|c}
\toprule
Method & \multicolumn{2}{c}{\ mIOU(\%)}\\
\hline
Outputsize &1/16& 1/8\\
\hline
\hline
 \noalign{\smallskip}
%  Res-101$^-$ & 69.26 \\
Res-50 & 69.15& 70.83\\
Res-50+GC & 71.24& 72.77\\
\hline
 \noalign{\smallskip}
Res-50+GCM ( $\delta =2$)& 72.36 & 74.21\\
Res-50+GCM ( $\delta =5$)& \textbf{72.45}&\textbf{74.50}\\
Res-50+GCM ( $\delta =10$)& 71.87& 74.30\\
\hline
\bottomrule
\end{tabular}
\end{center}
\caption{Ablation experiments of Global Context Module on Cityscapes validation set, $\delta$ denotes the amplitude of difference distribution of the global gated coefficient.}
\label{globalT}
\end{table}

\begin{figure*}[!t]
        \centering
        \includegraphics[width=1\linewidth]{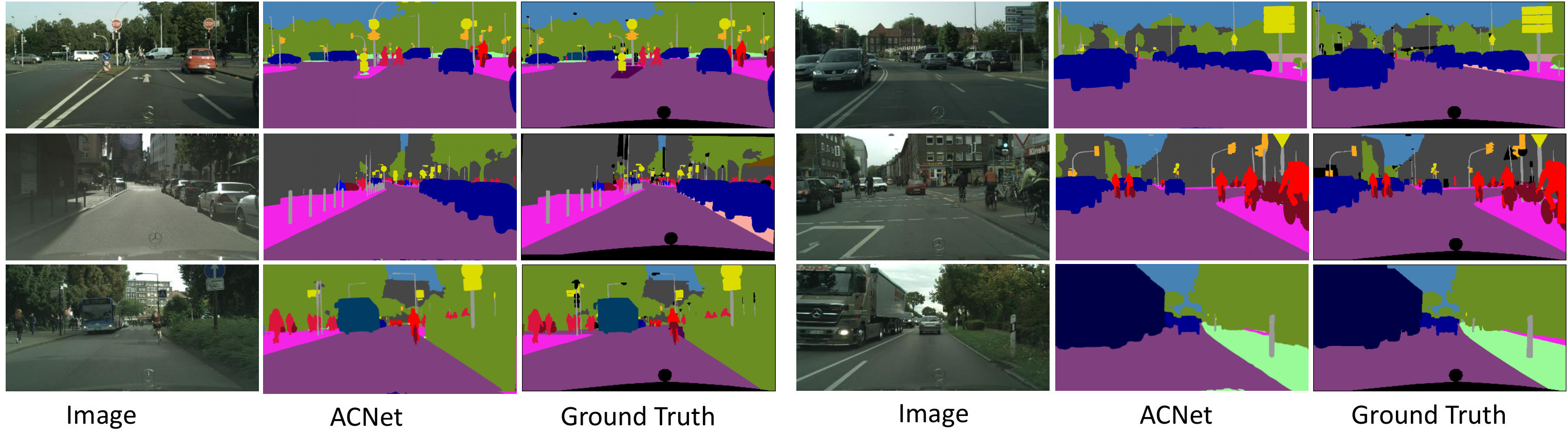}
  \caption{Example results of ACNet on Cityscapes validation set. (Best viewed in color)}
         \label{SPAM}
        \vspace{-1em}    
  
\end{figure*}

\begin{table}[t]
\begin{center}
\begin{tabular}{l|c}
\toprule
Method & mIOU(\%)\\
\hline
\hline
 \noalign{\smallskip}
Res-50 & 69.15\\
Res-50+GCM & 72.45\\
Res-50+GCM+LC & 73.48\\
\hline
 \noalign{\smallskip}
Res-50+GCM+LCM(1) & 74.03\\
Res-50+GCM+LCM(2)& 74.56\\ 
Res-50+GCM+LCM(3)& \textbf{74.67}\\
\hline
\bottomrule
\end{tabular}
\end{center}
\caption{Ablation experiments of Local Context Module on Cityscapes validation set, (n)  denotes the n times of fusion of local gated features in the LCM.  }
\label{localT}
     \vspace{-1em}    
\end{table}

\begin{table}[t]
\begin{center}
\begin{tabular}{l|c}
\toprule
Method & mIoU(\%)\\
\hline
\hline
 \noalign{\smallskip}
Res-50+ACB\#1 & 74.67\\
Res-50+ACB\#2 & 75.98\\
Res-50+ACB\#3& \textbf{76.53}\\ 
\hline
 \noalign{\smallskip}
Res-101+ACB\#3 & 77.42\\
Res-101+ACB\#3+MG & 78.50\\ 
Res-101+ACB\#3+MG+DA & 80.09\\
Res-101+ACB\#3+MG+DA+OHEM &80.89\\
Res-101+ACB\#3+MG+DA+OHEM+MS &\textbf{82.00}\\
\hline

\bottomrule
\end{tabular}
\end{center}
\caption{Ablation experiments of Adaptive Context Block on Cityscapes validation set, \#n  denotes the number of Adaptive Context Block, MG denotes multi-grid dilated convolution, DA denotes data augmentation with multi-scale input during training phase, MS denotes multi-scale testing. }
\vspace{-1.2em}
\label{ACBT}
\end{table}

\begin{table*}[tp]
\renewcommand\arraystretch{1.2}
    \footnotesize
    %\scriptsize
    %\tiny
    \setlength{\tabcolsep}{4pt}
    \begin{center}
    \begin{adjustbox}{max width=\textwidth}
        \begin{tabular}{ l | c |c c c c c c c c c c c c c c c c c c c  c}
            %\hline
            \toprule[1pt]
       Methods &  \rotatebox{90}{mIoU} &  \rotatebox{90}{road} &  \rotatebox{90}{sidewalk} &  \rotatebox{90}{building} & \rotatebox{90}{ wall} &  \rotatebox{90}{fence} &  \rotatebox{90}{pole} & \rotatebox{90}{traffic light} &  \rotatebox{90}{traffic sign}&  \rotatebox{90}{vegetation} &  \rotatebox{90}{terrain} &  \rotatebox{90}{sky} & \rotatebox{90}{person} &  \rotatebox{90}{rider} & \rotatebox{90}{car} &  \rotatebox{90}{truck}& \rotatebox{90}{ bus}& \rotatebox{90}{ train}& \rotatebox{90}{ motorcycle}&  \rotatebox{90}{bicycle}\\
      \hline
      \hline
       RefineNet~\cite{refinenet} & 73.6 & 98.2 & 83.3 & 91.3 & 47.8 & 50.4 & 56.1 & 66.9 & 71.3 & 92.3 & 70.3 & 94.8 & 80.9 & 63.3 & 94.5 & 64.6 & 76.1 & 64.3 & 62.2 & 70 \\
       DUC~\cite{wang2017understanding} & 77.6 & 98.5 & 85.5 & 92.8 & 58.6 & 55.5 & 65 & 73.5 & 77.9 & 93.3 & 72 & 95.2 & 84.8 & 68.5 & 95.4 & 70.9 & 78.8 & 68.7 & 65.9 & 73.8 \\  
       ResNet-38~\cite{wu2016wider} & 78.4 & 98.5 & 85.7 & 93.1 & 55.5 & 59.1 & 67.1 & 74.8 & 78.7 & 93.7 & 72.6 & 95.5 & 86.6 & 69.2 & 95.7 & 64.5 & 78.8 & 74.1 & 69 & 76.7 \\
    PSPNet~\cite{pspnet} & 78.4 & - & - & - & - & - & - & - & - & - & - & - & - & - & - & - & - & - & - & -  \\
       BiSeNet~\cite{yu2018bisenet} & 78.9 & - & - & - & - & - & - & - & - & - & - & - & - & - & - & - & - & - & - & -  \\ 
       PSANet~\cite{zhao2018psanet} & 80.1 & - & - & - & - & - & - & - & - & - & - & - & - & - & - & - & - & - & - & -  \\ 
       DenseASPP~\cite{yang2018denseaspp} & 80.6 & 98.7 & 87.1 & 93.4 & 60.7 & 62.7 & 65.6 & 74.6 & 78.5 & 93.6 & 72.5 & 95.4 & 86.2 & 71.9 & 96.0 & 78.0 & 90.3 & 80.7 & 69.7 & 76.8 \\  
       CCNet~\cite{huang2018ccnet} & 81.4 & - & - & - & - & - & - & - & - & - & - & - & - & - & - & - & - & - & - & -  \\ 
       DANet \cite{fu2018dual} & 81.5& 98.6 & 86.1 &93.5 & 56.1 & 63.3 & 69.7 & 77.3 & 81.3 & 93.9&72.9& 95.7 & 87.3 & 72.9& 96.2 & 76.8 & 89.4 & 86.5 & 72.2& 78.2\\  
 \hline 
       ACNet  & \textbf{82.3} & 98.7 & 87.1 & 93.9 & 61.6 & 61.8 &71.4& 78.7 & 81.7 & 94.0 &73.3&96.0& 88.5 &74.9 &96.5 & 77.1 & 89.0 & 89.2 & 71.4 & 79.0\\  
       \hline   
        \bottomrule[1pt]
        \end{tabular}
    \end{adjustbox}
    \end{center}
    \vspace{-1em}
\caption{ Category-wise comparison with state-of-the-art methods on Cityscapes testing set.  }

\vspace{-1.2em}
\label{cityset}
\end{table*}

\noindent{\bfseries Local Context Module:} We also propose  a local context module to refine spatial details. Since we need to generate the local gated coefficient for each pixel by inversing global gated coefficien, the local context module is built on the global context module.
Specifically, experiments are conducted on a dilated ResNet-50 with a GCM, then we  cascade local features from the outputs of ResNet block-2 with (LCM) and without (LC) local gated coefficient .

Results are shown in Table \ref{localT},  we can see that the local context improves the performance from 72.45\% to 73.48\%. When we adopt the  local gated coefficient to selectively fuse local context for each pixel once, the  performance is further improved to 74.03\%.
Reusing local gated feature brings  continuous improvements of the performance from 74.03\% to 74.67\%.

\noindent{\bfseries Adaptive Context Block:}  We further build an  adaptive context block and cascade it for three times to  obtain  high resolution predictions.  Results are listed in Table \ref{ACBT}. When we employ three  adaptive context blocks (ACB\#3), the performance is improved to 76.53\%, which verifies the effectiveness of our method. 

In addition, we visualize the  global gated coefficients in three  adaptive context blocks with different resolutions as shown in Figure \ref{attencity}. The images are from the validation set of Cityscapes.  We can find that the pixels with large  global gated coefficient perfer to dominant stuff and large objects,  such as the ``road'' in the first row and ``car'' in the last two rows.  These stuff and objects are improved in our method. In addition, the pixels with small  global gated coefficient perfer to small objects and edges, such as the ``traffic sign'' and ``pole'',``person'', etc. These spatial details are also be refined in our results.  A similar trend is also spotted in other images.

\noindent{\bfseries Some improvement strategies:} we follow the common procedure of \cite{deeplabv3,  fu2018dual,he2019adaptive,ding2019semantic,ding2019boundary,he2019dynamic} to further  improve the performance of ACNet: (1) A deeper and powerful network ResNet-101. (2) MG: Different dilated rates (4,8,16) in the last ResNet block.    (3) DA: We transform the input images with random scales (from 0.5 to 2.2) during training phase. (4) OHEM: The online hard example mining is also adopted. (5) MS:  we apply the multi-scale inputs with scales \{0.5 0.75, 1, 1.25, 1.5, 1.75, 2, 2.25\} as well as their mirrors for inference. 

Experimental results are shown in Table \ref{ACBT}, when employing a deeper backbone (ResNet101), ACNet obtains  77.42\% in terms of mean IoU. Then multi-grid dilated convolutional improves the performance by 1.08\%. Data augmentation with multi-scale input (DA) brings another 1.59\% improvement. OHEM increases the performance to  80.89\%. Finally, using multi-scale testing, we attains the best performance of 82.00\% on the validation set.

\noindent{\bfseries Compared with state-of-art methods:} We also compare our method with state-of-the-art methods on Cityscapes test set. Specifically, we fine tune our best model of ACNet
with only fine annotated trainval data, and submit our test results to the official
evaluation server.  For each method, we report the accuracy for each class and
the average class accuracy, which are reported in the original paper. Results are shown in Table. \ref{cityset}. We can see that our ACNet achieve a new state-of-the-art performance of 82.3\% \% $\footnote{\url{https://www.cityscapes-dataset.com/anonymous-results/?id=205de9c8857dbbe1ab61e7f5cc08be5ea290ece3c5125e3135d081139720d8f3}}$ on the test set. 
With the same backbone ResNet-101, our model outperforms DANet\cite{fu2018dual}. Moreover, ACNet also surpasses DenseASPP \cite{yang2018denseaspp} , which uses more powerful pretrained models, and is heigher than Deeplabv3+ \cite{chen2018encoder} (82.1\%), which uses extra the coarse annotations in training phase.

\subsection{Results on ADE20K dataset}

\begin{table}[t]
\begin{center}
\begin{tabular}{l|l |c| c}
\toprule
Backbone & Method & mIoU (\%)& PixAcc\%\\
\hline
\hline
% \noalign{\smallskip}
\multirow{5}{*}{Res-50}&Dilated FCN & 37.32 & 77.78 \\
&EncNet\cite{encnet}  & 41.11& 79.73 \\
&GCU\cite{li2018beyond}&42.60&79.51\\
&PSPNet\cite{pspnet}  & 42.78 & 80.76 \\
&PASNet\cite{zhao2018psanet}  & 42.98 & 80.92 \\
&ACNet  & \textbf{43.01} & \textbf{81.01} \\
\hline
% \noalign{\smallskip}
\multirow{7}{*}{Res-101}&UperNet\cite{xiao2018unified} &42.66&81.01\\
&PSPNet\cite{pspnet}  & 43.29 & 81.39\\
&DSSPN\cite{liang2018dynamic}  & 43.68 & 81.13 \\
&PASNet\cite{zhao2018psanet}  & 43.77& 81.51\\
&SGR \cite{liang2018symbolic}&44.32&81.43\\
&EncNet\cite{encnet}  & 44.65 & 81.19\\
&GCU\cite{li2018beyond}&44.81&81.19\\
&ACNet  & \textbf{45.90} & \textbf{81.96} \\
\hline
\bottomrule
\end{tabular}
\end{center}
\caption{Results of semantic segmentation on
ADE20K validation set.  }
\vspace{-1.2em}
\label{ADE20kv}
\end{table}

\begin{table}[t]
\begin{center}
\begin{tabular}{l| c}
\toprule
Method & Final score(\%)\\
\hline
\hline
 \noalign{\smallskip}
PSPNet269 (1st in place 2016)& 55.38\\
PSANet-101\cite{zhao2018psanet} & 55.46\\
CASIA\_IVA\_JD (1st in place 2017)& 55.47\\
EncNet-101 \cite{encnet} & 55.67\\
ACNet-101 &  \textbf{55.84}\\
\hline
\bottomrule
\end{tabular}
\end{center}
\caption{Results of semantic segmentation on
ADE20K testing set.  }
\vspace{-1.2em}
\label{ADE20kt}
\end{table}

In this subsection, we conduct experiments on the ADE20K dataset to
validate the effectiveness of our method.  Following previous works \cite{huang2018ccnet, li2018beyond,encnet,pspnet, zhao2018psanet}, data augmentation with multi-scale input and multi-scale testing are used.
We evalute ACNet by pixel-wise accuracy (PixelAcc) and mean of
class-wise intersection over union (mIoU).
Quantitative results are shown in Table.\ref{ADE20kv}. With ResNet50, the dilated FCN obtains 37.32\%/77.78\% in terms of mIoU and PixelAcc. When adopting our method, the performance is  improved by 5.69\%/3.23\%. When employing a deeper backbone ResNet101, ACNet achieves a new state-of-the-art performance
of 45.90\%/81.96\%, which outperforms the previous state-of-the-art methods. In addition,  we also fine tune our best model of ACNet-101 with trainval data, and submit our test results on the test set. The with single model of ACNet-101 gets final score as 55.84\%.
Among the approaches, most of methods \cite{pspnet,encnet,li2018beyond,zhang2017scale,zhao2018psanet,huang2018ccnet} attemp to  explore the global information by  aggregation  variant and relationship of the feature on the the top of the backbones. While our method focuses on  capturing the pixel-aware contexts from high and low-level features and achieves better performance.

%-------------------------------------------------------------------------

\begin{table}[t]
\begin{center}
%\resizebox{1\linewidth}{!}
  \begin{tabular} {l |l | c }
    \toprule[1pt]
    Backbone & Method & mIoU (\%)\\
    \hline \hline
    \multirow{5}{*}{Res-101}

    &Ding et al.\cite{ding2018context} &  51.6 \\
   &EncNet  \cite{encnet} &  51.7 \\
  &SGR \cite{liang2018symbolic}&52.5\\
   &DANet  \cite{fu2018dual} &  52.6 \\
    &    ACNet & \textbf{54.1} \\
       \hline
 \multirow{2}{*}{Res-152}& RefineNet \cite{refinenet} &  47.3\\ 
&MSCI\cite{Lin_2018_ECCV} &50.3\\
       \hline
Xception-71& Tian et al.\cite{CVPR19Tian7} &  52.5 \\
       \hline
    \bottomrule[1pt]
 \end{tabular}
\end{center}
\caption{Segmentation results on PASCAL Context  testing set.}
\vspace{-1em}
\label{pcontext}
\end{table}

\subsection{Results on PASCAL Context  Dataset }

We also carry out experiments on the PASCAL Context dataset to further demonstrate the effectiveness of  ACNet.  We employ the ACNet-101 network with the same
training strategy on ADE20K and compare our model with previous state-of-the-art methods.
The results are  reported in Table \ref{pcontext}. ACNet obtains a Mean IoU of 54.1\%, which surpasses previous published methods.
Among the approaches, the recent methods\cite{Lin_2018_ECCV,CVPR19Tian7} use more powerful network(e.g. ResNet-152 and Xception-71) as encoder network  and  fuse high-and low-level feature in decoder network, our method outperforms them  by  a relatively large margin. 
%-------------------------------------------------------------------------

\subsection{Results on  COCO stuff Dataset }

Finally, we  demonstrate the effectiveness of  ACNet on the COCO stuff dataset.  The ACNet-101 network is also employed. The COCO stuff  results are  reported in Table \ref{cocostuff}. ACNet  achieves performance of 40.1\% Mean IoU, which also outperforms other state-of-the-art methods. 
 
\begin{table}[t]
\begin{center}
%\resizebox{1\linewidth}{!}
  \begin{tabular} {l |l | c }
    \toprule[1pt]
    Backbone & Method & mIoU(\%)\\
    \hline \hline
    \multirow{5}{*}{Res-101}
    &RefineNet  \cite{refinenet} &  33.6 \\
    &Ding et al.\cite{ding2018context} &  35.7 \\
   &DSSPN\cite{liang2018dynamic}  &  38.9 \\
 &SGR \cite{liang2018symbolic}&39.1\\
   &DANet\cite{fu2018dual} & 39.7 \\
    &    ACNet & \textbf{40.1} \\
       \hline
    \bottomrule[1pt]
 \end{tabular}
\end{center}
\caption{Segmentation results on COCO Stuff testing set.}
\vspace{-1em}
\label{cocostuff}
\end{table}

\section{Conclusion}
In this paper, we present a novel network of ACNet to capture pixel-aware adaptive contexts for scene parsing, in which a global context module and a local context module are carefully designed and jointly employed as an adaptive context block to obtain a competitive fusion of the both contexts for each position. Our work is motivated by the observation that the global context from high-level features helps the categorization of some large semantic confused regions, while the local context from lower-level visual features helps to generate sharp boundaries or clear details. 
Extensive experiments demonstrate the outstanding performance of ACNet compared with other state-of-the-art methods. We believe such an adaptive context block can also be extended to other vision applications including object detection, pose estimation, and fine-grained recognition.

\noindent\textbf{Acknowledgement:} This work was supported by National Natural Science Foundation of China (61872366 and 61872364) and Beijing Natural Science Foundation (4192059)
{\small
\bibliographystyle{ieee_fullname}
\bibliography{references}

\begin{thebibliography}{10}\itemsep=-1pt

\bibitem{caesar2018coco}
Holger Caesar, Jasper Uijlings, and Vittorio Ferrari.
\newblock Coco-stuff: Thing and stuff classes in context.
\newblock In {\em Proceedings of the IEEE Conference on Computer Vision and
  Pattern Recognition}, pages 1209--1218, 2018.

\bibitem{deeplabv2}
Liang{-}Chieh Chen, George Papandreou, Iasonas Kokkinos, Kevin Murphy, and
  Alan~L. Yuille.
\newblock Deeplab: Semantic image segmentation with deep convolutional nets,
  atrous convolution, and fully connected crfs.
\newblock {\em IEEE Transactions on Pattern Analysis and Machine
  Intelligence.}, 40(4):834--848, 2018.

\bibitem{deeplabv3}
Liang{-}Chieh Chen, George Papandreou, Florian Schroff, and Hartwig Adam.
\newblock Rethinking atrous convolution for semantic image segmentation.
\newblock {\em CoRR}, abs/1706.05587, 2017.

\bibitem{chen2018encoder}
Liang-Chieh Chen, Yukun Zhu, George Papandreou, Florian Schroff, and Hartwig
  Adam.
\newblock Encoder-decoder with atrous separable convolution for semantic image
  segmentation.
\newblock In {\em Proceedings of the European Conference on Computer Vision
  (ECCV)}, pages 801--818, 2018.

\bibitem{cityscapes}
Marius Cordts, Mohamed Omran, Sebastian Ramos, Timo Rehfeld, Markus Enzweiler,
  Rodrigo Benenson, Uwe Franke, Stefan Roth, and Bernt Schiele.
\newblock The cityscapes dataset for semantic urban scene understanding.
\newblock In {\em {IEEE} Conference on Computer Vision and Pattern
  Recognition}, pages 3213--3223, 2016.

\bibitem{ding2019boundary}
Henghui Ding, Xudong Jiang, Ai~Qun Liu, Nadia~Magnenat Thalmann, and Gang Wang.
\newblock Boundary-aware feature propagation for scene segmentation.
\newblock In {\em Proceedings of the IEEE International Conference on Computer
  Vision}, 2019.

\bibitem{ding2018context}
Henghui Ding, Xudong Jiang, Bing Shuai, Ai~Qun Liu, and Gang Wang.
\newblock Context contrasted feature and gated multi-scale aggregation for
  scene segmentation.
\newblock In {\em Proceedings of the IEEE Conference on Computer Vision and
  Pattern Recognition}, pages 2393--2402, 2018.

\bibitem{ding2019semantic}
Henghui Ding, Xudong Jiang, Bing Shuai, Ai~Qun Liu, and Gang Wang.
\newblock Semantic correlation promoted shape-variant context for segmentation.
\newblock In {\em Proceedings of the IEEE Conference on Computer Vision and
  Pattern Recognition}, pages 8885--8894, 2019.

\bibitem{fu2017stacked}
Jun Fu, Jing Liu, Yuhang Wang, and Hanqing Lu.
\newblock Stacked deconvolutional network for semantic segmentation.
\newblock {\em arXiv preprint arXiv:1708.04943}, 2017.

\bibitem{ghiasi2016laplacian}
Golnaz Ghiasi and Charless~C. Fowlkes.
\newblock Laplacian pyramid reconstruction and refinement for semantic
  segmentation.
\newblock In {\em the European Conference on Computer Vision}, pages 519--534,
  2016.

\bibitem{he2019dynamic}
Junjun He, Zhongying Deng, and Yu Qiao.
\newblock Dynamic multi-scale filters for semantic segmentation.
\newblock In {\em Proceedings of the International Conference on Computer
  Vision}, 2019.

\bibitem{he2019adaptive}
Junjun He, Zhongying Deng, Lei Zhou, Yali Wang, and Yu Qiao.
\newblock Adaptive pyramid context network for semantic segmentation.
\newblock In {\em Proceedings of the IEEE Conference on Computer Vision and
  Pattern Recognition}, pages 7519--7528, 2019.

\bibitem{he2016deep}
Kaiming He, Xiangyu Zhang, Shaoqing Ren, and Jian Sun.
\newblock Deep residual learning for image recognition.
\newblock In {\em Proceedings of the IEEE conference on computer vision and
  pattern recognition}, pages 770--778, 2016.

\bibitem{huang2018ccnet}
Zilong Huang, Xinggang Wang, Lichao Huang, Chang Huang, Yunchao Wei, and Wenyu
  Liu.
\newblock Ccnet: Criss-cross attention for semantic segmentation.
\newblock {\em arXiv preprint arXiv:1811.11721}, 2018.

\bibitem{hung2017scene}
Wei-Chih Hung, Yi-Hsuan Tsai, Xiaohui Shen, Zhe Lin, Kalyan Sunkavalli, Xin Lu,
  and Ming-Hsuan Yang.
\newblock Scene parsing with global context embedding.
\newblock In {\em Proceedings of the IEEE International Conference on Computer
  Vision}, pages 2631--2639, 2017.

\bibitem{fu2018dual}
Fu Jun, Liu Jing, Tian Haijie, Li Yong, Bao Yongjun, Fang Zhiwei, and Lu
  Hanqing.
\newblock Dual attention network for scene segmentation.
\newblock In {\em Proceedings of thr IEEE Conference on Computer Vision and
  Pattern Recognition}, 2019.

\bibitem{li2018pyramid}
Hanchao Li, Pengfei Xiong, Jie An, and Lingxue Wang.
\newblock Pyramid attention network for semantic segmentation.
\newblock {\em arXiv preprint arXiv:1805.10180}, 2018.

\bibitem{li2018beyond}
Yin Li and Abhinav Gupta.
\newblock Beyond grids: Learning graph representations for visual recognition.
\newblock In {\em Advances in Neural Information Processing Systems}, pages
  9245--9255, 2018.

\bibitem{liang2018symbolic}
Xiaodan Liang, Zhiting Hu, Hao Zhang, Liang Lin, and Eric~P Xing.
\newblock Symbolic graph reasoning meets convolutions.
\newblock In {\em Advances in Neural Information Processing Systems}, pages
  1858--1868, 2018.

\bibitem{liang2018dynamic}
Xiaodan Liang, Hongfei Zhou, and Eric Xing.
\newblock Dynamic-structured semantic propagation network.
\newblock In {\em Proceedings of the IEEE Conference on Computer Vision and
  Pattern Recognition}, pages 752--761, 2018.

\bibitem{Lin_2018_ECCV}
Di Lin, Yuanfeng Ji, Dani Lischinski, Daniel Cohen-Or, and Hui Huang.
\newblock Multi-scale context intertwining for semantic segmentation.
\newblock In {\em The European Conference on Computer Vision (ECCV)}, September
  2018.

\bibitem{refinenet}
Guosheng Lin, Anton Milan, Chunhua Shen, and Ian~D. Reid.
\newblock Refinenet: Multi-path refinement networks for high-resolution
  semantic segmentation.
\newblock In {\em {IEEE} Conference on Computer Vision and Pattern
  Recognition}, pages 5168--5177, 2017.

\bibitem{parsenet}
Wei Liu, Andrew Rabinovich, and Alexander~C. Berg.
\newblock Parsenet: Looking wider to see better.
\newblock In {\em the International Conference on Learning Representations},
  2016.

\bibitem{FCN}
Jonathan Long, Evan Shelhamer, and Trevor Darrell.
\newblock Fully convolutional networks for semantic segmentation.
\newblock In {\em Proceedings of the IEEE conference on computer vision and
  pattern recognition}, pages 3431--3440, 2015.

\bibitem{luo2016understanding}
Wenjie Luo, Yujia Li, Raquel Urtasun, and Richard Zemel.
\newblock Understanding the effective receptive field in deep convolutional
  neural networks.
\newblock In {\em Advances in neural information processing systems}, pages
  4898--4906, 2016.

\bibitem{pcontext}
Roozbeh Mottaghi, Xianjie Chen, Xiaobai Liu, Nam{-}Gyu Cho, Seong{-}Whan Lee,
  Sanja Fidler, Raquel Urtasun, and Alan~L. Yuille.
\newblock The role of context for object detection and semantic segmentation in
  the wild.
\newblock In {\em 2014 {IEEE} Conference on Computer Vision and Pattern
  Recognition}, pages 891--898, 2014.

\bibitem{DAGRNN}
Bing Shuai, Zhen Zuo, Bing Wang, and Gang Wang.
\newblock Scene segmentation with dag-recurrent neural networks.
\newblock {\em {IEEE} Trans. Pattern Anal. Mach. Intell.}, pages 1480--1493,
  2018.

\bibitem{CVPR19Tian7}
Zhi Tian, Tong He, Chunhua Shen, and Youliang Yan.
\newblock Decoders matter for semantic segmentation: Data-dependent decoding
  enables flexible feature aggregation.
\newblock In {\em IEEE Conference on Computer Vision and Pattern Recognition
  (CVPR'19)}, 2019.

\bibitem{wang2017understanding}
Panqu Wang, Pengfei Chen, Ye Yuan, Ding Liu, Zehua Huang, Xiaodi Hou, and
  Garrison~W. Cottrell.
\newblock Understanding convolution for semantic segmentation.
\newblock In {\em {IEEE} Winter Conference on Applications of Computer Vision},
  pages 1451--1460, 2018.

\bibitem{wu2016wider}
Zifeng Wu, Chunhua Shen, and Anton van~den Hengel.
\newblock Wider or deeper: Revisiting the resnet model for visual recognition.
\newblock {\em arXiv preprint arXiv:1611.10080}, 2016.

\bibitem{xiao2018unified}
Tete Xiao, Yingcheng Liu, Bolei Zhou, Yuning Jiang, and Jian Sun.
\newblock Unified perceptual parsing for scene understanding.
\newblock In {\em Proceedings of the European Conference on Computer Vision
  (ECCV)}, pages 418--434, 2018.

\bibitem{yang2018denseaspp}
Maoke Yang, Kun Yu, Chi Zhang, Zhiwei Li, and Kuiyuan Yang.
\newblock Denseaspp for semantic segmentation in street scenes.
\newblock In {\em Proceedings of the IEEE Conference on Computer Vision and
  Pattern Recognition}, pages 3684--3692, 2018.

\bibitem{yu2018bisenet}
Changqian Yu, Jingbo Wang, Chao Peng, Changxin Gao, Gang Yu, and Nong Sang.
\newblock Bisenet: Bilateral segmentation network for real-time semantic
  segmentation.
\newblock In {\em Proceedings of the European Conference on Computer Vision
  (ECCV)}, pages 325--341, 2018.

\bibitem{yu2018learning}
Changqian Yu, Jingbo Wang, Chao Peng, Changxin Gao, Gang Yu, and Nong Sang.
\newblock Learning a discriminative feature network for semantic segmentation.
\newblock In {\em {CVPR}}, 2018.

\bibitem{yu2015multi}
Fisher Yu and Vladlen Koltun.
\newblock Multi-scale context aggregation by dilated convolutions.
\newblock In {\em the International Conference on Learning Representations},
  2016.

\bibitem{yuan2018ocnet}
Yuhui Yuan and Jingdong Wang.
\newblock Ocnet: Object context network for scene parsing.
\newblock {\em arXiv preprint arXiv:1809.00916}, 2018.

\bibitem{encnet}
Hang Zhang, Kristin Dana, Jianping Shi, Zhongyue Zhang, Xiaogang Wang, Ambrish
  Tyagi, and Amit Agrawal.
\newblock Context encoding for semantic segmentation.
\newblock In {\em The IEEE Conference on Computer Vision and Pattern
  Recognition (CVPR)}, 2018.

\bibitem{zhang2017scale}
Rui Zhang, Sheng Tang, Yongdong Zhang, Jintao Li, and Shuicheng Yan.
\newblock Scale-adaptive convolutions for scene parsing.
\newblock In {\em Proceedings of the IEEE International Conference on Computer
  Vision}, pages 2031--2039, 2017.

\bibitem{exfuse}
Zhenli Zhang, Xiangyu Zhang, Chao Peng, Xiangyang Xue, and Jian Sun.
\newblock Exfuse: Enhancing feature fusion for semantic segmentation.
\newblock In Vittorio Ferrari, Martial Hebert, Cristian Sminchisescu, and Yair
  Weiss, editors, {\em Computer Vision -- ECCV 2018}, pages 273--288, 2018.

\bibitem{pspnet}
Hengshuang Zhao, Jianping Shi, Xiaojuan Qi, Xiaogang Wang, and Jiaya Jia.
\newblock Pyramid scene parsing network.
\newblock In {\em {IEEE} Conference on Computer Vision and Pattern
  Recognition}, pages 6230--6239, 2017.

\bibitem{zhao2018psanet}
Hengshuang Zhao, Yi Zhang, Shu Liu, Jianping Shi, Chen Change~Loy, Dahua Lin,
  and Jiaya Jia.
\newblock Psanet: Point-wise spatial attention network for scene parsing.
\newblock In {\em Proceedings of the European Conference on Computer Vision
  (ECCV)}, pages 267--283, 2018.

\bibitem{zhou2017scene}
Bolei Zhou, Hang Zhao, Xavier Puig, Sanja Fidler, Adela Barriuso, and Antonio
  Torralba.
\newblock Scene parsing through {ADE20K} dataset.
\newblock In {\em {IEEE} Conference on Computer Vision and Pattern
  Recognition}, pages 5122--5130, 2017.

\end{thebibliography}
}

\end{document}